\documentclass[journal]{IEEEtran}

\usepackage{times}
\usepackage{epsfig}
\usepackage{graphicx}
\usepackage{amsmath}
\usepackage{amssymb}
\usepackage{multirow}
\usepackage{cite}
\usepackage{comment}
\usepackage{algorithm,algpseudocode}
\usepackage{amsmath}
\usepackage{amsfonts}
\usepackage{amssymb}
\usepackage{multirow}

\usepackage{booktabs}

\usepackage{tikz}
\ifCLASSINFOpdf

\else

\fi

\begin{document}

\title{Hierarchical Spatiotemporal Transformers for Video Object Segmentation}

\author{Jun-Sang Yoo,
        Hongjae Lee,
        and~Seung-Won Jung, \emph{Senior Member, IEEE}
\thanks{\textit{Corresponding author: Seung-Won Jung.}}
\thanks{J.-S. Yoo, H. Lee, and S.-W. Jung are with the Department
of Electrical and Electronic Engineering, Korea University, Seoul,
Korea. (e-mail: junsang7777@naver.com; jimmy9704@korea.ac.kr; swjung83@korea.ac.kr)}
}

\maketitle

\begin{abstract}
   This paper presents a novel framework called HST for semi-supervised video object segmentation (VOS). HST extracts image and video features using the latest Swin Transformer and Video Swin Transformer to inherit their inductive bias for the spatiotemporal locality, which is essential for temporally coherent VOS. To take full advantage of the image and video features, HST casts image and video features as a query and memory, respectively. By applying efficient memory read operations at multiple scales, HST produces hierarchical features for the precise reconstruction of object masks. HST shows effectiveness and robustness in handling challenging scenarios with occluded and fast-moving objects under cluttered backgrounds. In particular, HST-B outperforms the state-of-the-art competitors on multiple popular benchmarks, i.e., YouTube-VOS ($85.0\%$), DAVIS 2017 ($85.9\%$), and DAVIS 2016 ($94.0\%$).
\end{abstract}

\begin{IEEEkeywords}
Video object segmentation; transformers; spatiotemporal features; hierarchical memory read
\end{IEEEkeywords}

\IEEEpeerreviewmaketitle

\section{Introduction}
\IEEEPARstart {S}{emi}-supervised video object segmentation (VOS) is the task of extracting a target object from a video sequence given an object mask of the first frame. It is a very challenging task because the appearance of the target object can change drastically over time. In addition, occlusion, cluttered backgrounds, and other objects similar to the target object make the task further challenging. Extensive research has been conducted on semi-supervised VOS over the last decade. The interested reader can refer to~\cite{yao2020video,gao2022deep,perazzi2016benchmark} for a systematic literature review.  

\begin{figure*}[ht]
  \centering
    \includegraphics[width=1.0\linewidth]{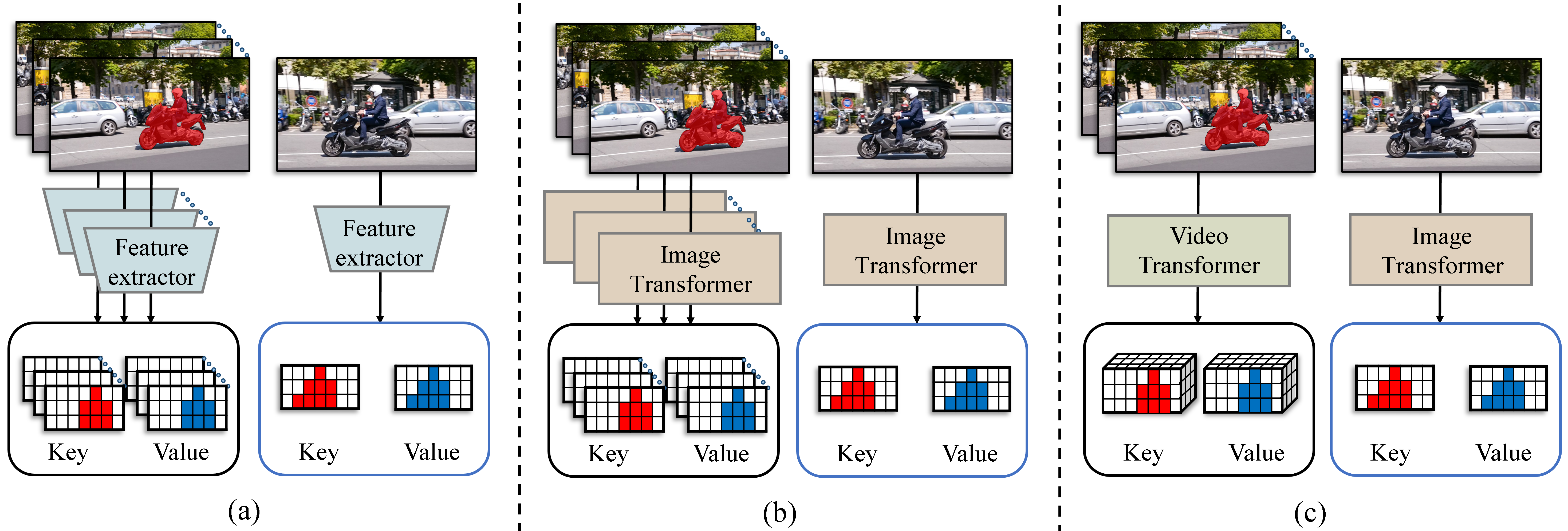}
  \caption{Comparison of the methods for extracting key and value maps from current and past frames: (a) Feature extractor is applied to each frame, (b) image Transformer is applied to each frame, and (c) image and video Transformers are applied to current and past frames, respectively.}
  \label{fig:fig1intro}
\end{figure*}

Recently, memory-based VOS methods~\cite{hu2018videomatch, oh2019video, lu2020video, huang2020fast, wang2019ranet, cheng2021modular, seong2020kernelized, liang2020video, li2020fast, hu2017maskrnn, hu2021learning, xie2021efficient, wang2021swiftnet, liu2022learning, park2022per,wen2020dmvos} have achieved remarkable performance. The key idea is to build a memory containing the information from the past frames with given or predicted masks and use the current frame as a query for matching. As shown in Figure~\ref{fig:fig1intro}(a), these methods typically apply a convolutional neural network (CNN)-based encoder to each frame and perform dense matching between the features extracted from the query and memory. Due to the non-local nature of this matching, they show robustness in handling moving objects and cameras. In particular, the space-time memory network (STM)~\cite{oh2019video} introduces a space-time memory read operation that performs dense matching between the query and the memory in the feature space to cover all space-time pixel locations. However, the global-to-global matching in STM requires high computational complexity and suffers from false matching to objects similar to the target object. Therefore, many follow-up studies attempted to enforce local constraints using kernelized memory~\cite{seong2020kernelized} and optical flow~\cite{xie2021efficient}. 

Meanwhile, the success of the Vision Transformer (ViT)~\cite{dosovitskiy2020image} has brought significant attention to a Transformer-based solution for VOS. Several recent Transformer-based methods~\cite{wang2021end, mei2021transvos, duke2021sstvos, yang2021associating} have shown state-of-the-art performance on several VOS benchmarks. However, these methods apply an image Transformer to each frame, as shown in Figure~\ref{fig:fig1intro}(b), and are thus still challenging to enforce Transformers to handle the temporal coherence of the segmentation. In this paper, we introduce a new approach that fully exploits spatiotemporal features for semi-supervised VOS. Inspired by the Swin Transformer~\cite{liu2021swin} and its extension to video frames, called the Video Swin Transformer~\cite{liu2022video}, we propose a novel integration of them for VOS, called HST. HST first extracts multi-scale features from the image and video using their respective Transformers, as shown in Figure~\ref{fig:fig1intro}(c). Then, the image features from the current frame are used as a query, and the video features from the past frames and their object masks are used as memory. Although HST performs dense matching between the query and memory, it does not suffer from false matching to objects similar to the target object due to the locality inductive bias of the Swin Transformers. We also apply an efficient hierarchical memory read operation to reduce computational complexity. HST shows robustness in segmenting small, fast-moving, and occluded objects under cluttered backgrounds. Our baseline model, HST-B, yields competitive performance in several VOS benchmarks, including the YouTube-VOS 2018 and 2019 validation datasets (85.0$\%$ \& 84.9$\%$) and the DAVIS 2016 validation (94.0$\%$) and 2017 validation and test datasets (85.9$\%$ \& 79.9$\%$). 

The main contributions are summarized as follows:

\begin{itemize}

    \item We propose a Swin Transformer-inspired VOS framework called HST that uses image and video Swin Transformers to extract spatial and spatiotemporal features. To the best of our knowledge, HST is the first to integrate image and video Swin Transformers for VOS.
    
    \item We apply a dedicated memory read operation for HST that efficiently measures the similarities between multi-scale spatial and spatiotemporal features.
    
    \item Experimental results on the DAVIS and YouTube-VOS datasets demonstrate the state-of-the-art performance of HST.

\end{itemize}

\section{Related Work}
\subsection{Semi-supervised Video Object Segmentation} Semi-supervised VOS methods have been developed to propagate the manual annotation from the first frame to the entire video sequence. Early semi-supervised VOS methods, such as OSVOS~\cite{caelles2017one} and MoNet~\cite{xiao2018monet}, fine-tune pre-trained networks at test time using the annotation from the first frame as the ground-truth. OnAVOS~\cite{voigtlaender2017online} applies an online adaptation mechanism to use pixels with very confident predictions from the following frames as additional training examples. MaskTrack~\cite{perazzi2017learning} and PReMVOS~\cite{luiten2018premvos} further estimate optical flow to facilitate the propagation of the segmentation mask.

Although promising results have been shown, online learning-based methods inevitably have high computational complexity, restricting their practical use. Recent efforts thus have been devoted to offline learning-based methods such that the trained networks can robustly handle any input videos without additional training. To this end, OSMN~\cite{yang2018efficient} uses spatial and visual modulators to adapt the segmentation model to the appearance of a specific object. VideoMatch~\cite{hu2018videomatch} applies a soft matching layer to compute the similarity of the foreground and background between the first frame and every input frame. FEELVOS~\cite{voigtlaender2019feelvos} and CFBI~\cite{yang2021collaborative,yang2020collaborative} perform pixel-level matching not only between the first and current frames but also between the previous and current frames. STM~\cite{oh2019video} embeds the past frames and their prediction masks in memory and uses the current frame as the query for global matching. KMN~\cite{seong2020kernelized}, RMNet~\cite{xie2021efficient}, and HMMN~\cite{seong2021hierarchical} further use local constraints such as optical flow and kernel to overcome the drawback of global matching. STCN~\cite{cheng2021rethinking} extracts key features for each image independently for effective feature reuse and replaces dot product by L2 similarity for better memory coverage.

\begin{figure*}[ht]
  \centering
    \includegraphics[width=1.0\linewidth]{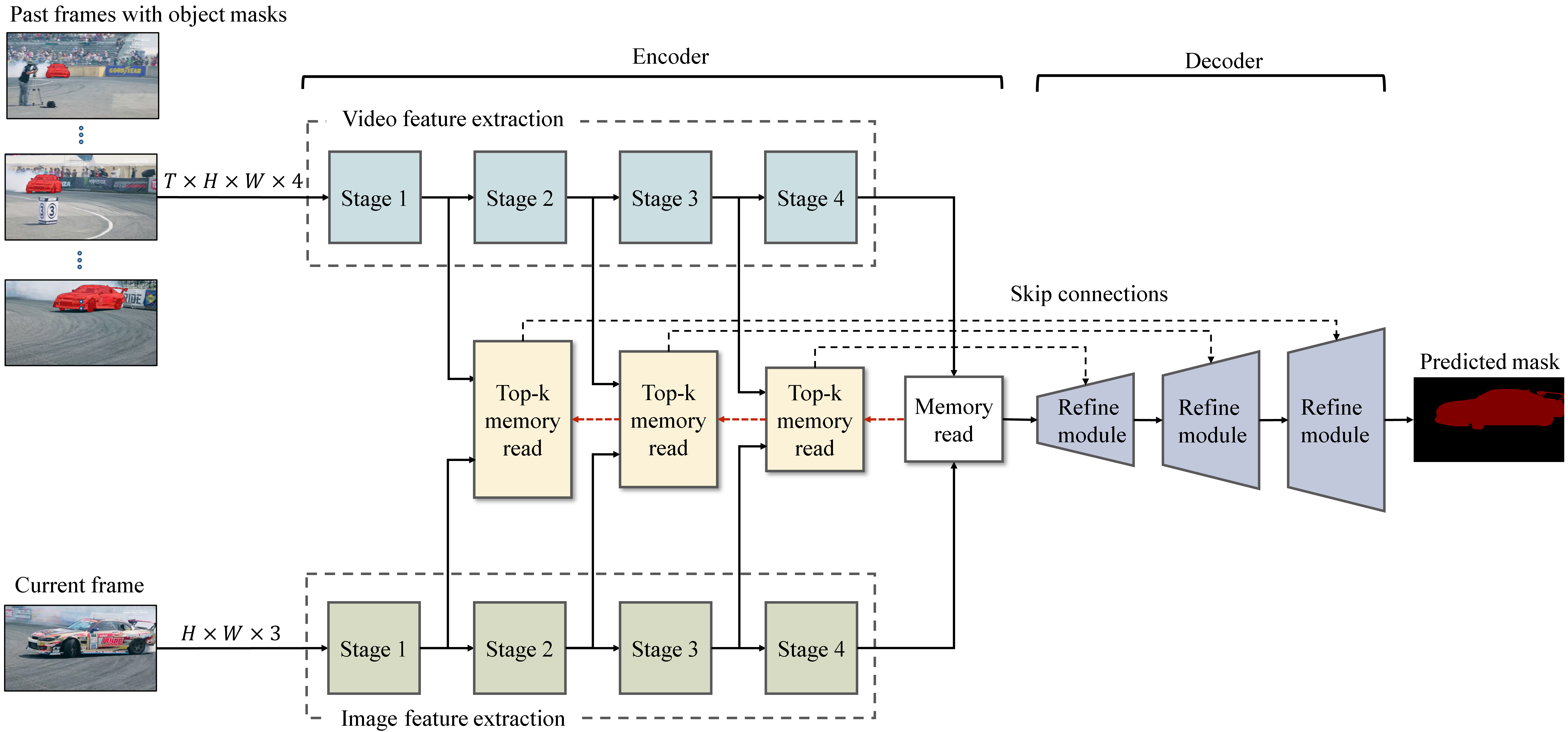}
  \caption{An overview of HST. The encoder consists of image and video Transformers to extract multi-scale spatial and spatiotemporal features from current and past frames. The memory read block performs dense matching on the coarsest scale, and the top-k memory read blocks operate on the finer scales. The decoder receives multi-scale features from the encoder and produces a final mask prediction.}
  \label{fig:fig2network}
\end{figure*}

\subsection{Vision Transformers}
The Transformer~\cite{vaswani2017attention} has been introduced as a network architecture solely based on attention mechanisms. Compared to recurrent neural networks (RNNs) that require extensive sequential operations, Transformer networks are more parallelizable and require less training time, making them attractive to several natural language processing tasks~\cite{bahdanau2014neural, sutskever2014sequence, cho2014learning, devlin2018bert}. Recently, Transformer networks have been successfully applied to many computer vision tasks and have shown significant performance improvement over CNN-based networks. The representative work called ViT~\cite{dosovitskiy2020image} divides an input image into non-overlapping patches and performs linear embedding to construct input for the Transformer encoder. DeiT~\cite{touvron2021training} integrates a teacher-student strategy to the Transformer such that the student model can be efficiently trained on a small dataset. A notable extension of ViT called Swin Transformer~\cite{liu2021swin} builds multi-resolution feature maps on Transformers and restricts self-attention within local windows, leading to linear computational complexity with respect to image size. Video Swin Transformer~\cite{liu2022video} expands the scope of local attention from the spatial domain to the spatiotemporal domain, achieving state-of-the-art accuracy on several video recognition benchmarks. Furthermore, the powerful feature representation ability of the Swin Transformer has reached outstanding performance on a variety of tasks, including video categorization and image inpainting~\cite{cai2022multiple,huang2022atrous, deng2021learning}.


\subsection{Transformer-based Segmentation}

Several recent endeavors have been made to apply vision Transformers to dense prediction tasks. DETR~\cite{carion2020end} integrates a CNN backbone with a Transformer encoder and decoder to build a fully end-to-end object detector and shows that dense prediction tasks such as panoptic segmentation can be handled by adding a mask head on top of the decoder outputs. 
TCTr~\cite{AZIERE2022104567} utilizes the temporal convolution transformer network, an integration of attention and depth-wise convolution, for action segmentation.
SegFormer~\cite{xie2021segformer} obtains a segmentation mask using a hierarchical pyramid ViT architecture as an encoder and a simple MLP-based structure with upsampling operations as a decoder. Segmenter~\cite{strudel2021segmenter} exploits a mask Transformer decoder to predict a better segmentation mask. Towards a more VOS-dedicated Transformer design, VIS~\cite{wang2021end} applies an instance sequence matching and segmentation strategy. TransVOS~\cite{mei2021transvos} extracts features from the current frame and reference sets and feeds them to the Transformer encoder to model the temporal and spatial relationships. SST~\cite{duke2021sstvos} uses a sparse attention-based Transformer block to extract pixel-level embedding and spatial-temporal features. 
AOT~\cite{yang2021associating} associates multiple target objects into the same embedding space to perform multi-object segmentation as efficiently as single object segmentation. AOT also shows that the performance can be further improved by changing a ResNet encoder to a Swin Transformer encoder.

However, many Transformer-based VOS methods still use a CNN-based encoder for feature extraction~\cite{duke2021sstvos,mei2021transvos,yang2021associating}, limiting the modeling capacity of Transformers. Fully Transformer-based feature extraction methods have been attempted, but they apply the standard Transformer or Swin Transformer~\cite{yang2021associating} to each frame separately, leading to sub-optimal extraction of spatiotemporal information in a video sequence. HST integrates image and video Transformers towards a complete spatiotemporal feature extraction for VOS.

\section{Approach}
We explain our method for segmenting one target object in a video, but multi-object segmentation can be readily conducted by following the common standard of independent segmentation and merging~\cite{oh2019video,seong2020kernelized,seong2021hierarchical,cheng2021rethinking}. We use the features extracted from the past frames (with given or estimated object masks) and the current frame as memory and query, respectively. The query should contain spatial information, such as the position, shape, and texture of the target object, and the memory should contain spatiotemporal information, such as the trajectory and deformation of the target object and changes in the background to support temporally coherent target object segmentation. To this end, we present HST that can fully exploit spatial and spatiotemporal information from the current and past frames. Moreover, since dense matching between the query and memory is needed to take full advantage of the information in video frames, we design a hierarchical memory read operation that efficiently matches multi-scale spatial and spatiotemporal features. 


Figure~\ref{fig:fig2network} illustrates the overall flow of HST. We adopt Swin Transformer~\cite{liu2021swin} and Video Swin Transformer~\cite{liu2022video} to design a query encoder (image as input) and a memory encoder (images and masks as input). For brevity, we call these two Swin Transformers image Transformer and video Transformer, respectively. Each Transformer extracts multi-scale features, resulting in the key and value maps for matching with each other. The decoder takes all the computed features and outputs a mask prediction. The following subsections detail each component of HST.

\subsection{Image Feature Extraction}
Our image feature extractor is based on Swin Transformer~\cite{liu2021swin} that incorporates inductive bias for the spatial locality, which is preferable for dense prediction tasks such as VOS. Image Transformer first splits a current frame of size ${H\times W\times 3}$ into non-overlapping patches of size $P_x\times P_y\times 3$ and applies a linear embedding layer, resulting in a $C$-dimensional embedding for each patch or ``token''. Unlike the standard Transformer that computes self-attention across all tokens~\cite{vaswani2017attention}, Swin Transformer computes self-attention only within each window. The query encoder of HST consists of four stacks of Swin Transformer blocks with patch merging blocks for generating multi-scale features~\cite{liu2021swin}. To introduce cross-window connections, a Swin Transformer block is embodied with consecutive multi-head self-attention units with and without a shifted window.

\subsection{Video Feature Extraction}
Our video feature extractor is based on Video Swin Transformer~\cite{liu2022video} that extends the scope of local attention from the spatial domain to the spatiotemporal domain. Specifically, the past $T$ frames and their corresponding object masks with size $T\times H\times W\times 4$ (RGB + mask) are divided into non-overlapping patches of size $P_t \times P_x\times P_y\times 4$, followed by a linear embedding layer to obtain a $C$-dimensional embedding for each token. To incorporate inductive bias for the spatiotemporal locality, video Transformer computes self-attention only within each 3D window. The memory encoder of HST consists of four stacks of Video Swin Transformer blocks with patch merging blocks for generating multi-scale features~\cite{liu2022video}, where each Video Swin Transformer block is embodied with consecutive multi-head self-attention units with and without a 3D shifted window.

\subsection{Memory Read and Decoding}
\subsubsection{Memory Read}
We now have image and video features ready to use for object segmentation. Let $F_{image}^{i} \in\mathbb{R}^{H_i \times W_i \times C_i}$ and $F_{video}^{i}\in\mathbb{R}^{T_i\times H_i \times W_i \times C_i}$ denote image and video features obtained after the $i$-th stage of the query encoder and memory encoder, respectively. The feature dimensions are given as ${H_i} = H \times {\left( {\frac{1}{2}} \right)^{i + 1}}$, ${W_i} = W \times {\left( {\frac{1}{2}} \right)^{i + 1}}$, and ${C_i} = C \times {2^{i - 1}}$~\cite{liu2021swin, liu2022video}. We fix ${T_i}$ to $T$ to maintain the temporal resolution. Considering $F_{image}^{i}$ as a query and $F_{video}^{i}$ as memory, we extract key and value maps from them~\cite{oh2019video}. The key and value maps of the query are denoted as $k_i^Q\in\mathbb{R}^{ \frac{C_i}{8} \times H_i W_i}$ and $v_i^Q\in\mathbb{R}^{\frac{C_i}{2}  \times H_i W_i}$, respectively, and those of the memory are denoted as $k_i^M\in\mathbb{R}^{\frac{C_i}{8} \times T_iH_i W_i  }$ and $v_i^M\in\mathbb{R}^{\frac{C_i}{2} \times T_iH_i W_i  }$, respectively. 

Due to extremely high dimensionality of the key and value maps, we apply dense matching between the query and memory only at the last stage as follows: 
\begin{equation}\label{eq:s4}
  s_4(\textbf{q}, \textbf{p}) =  \left(k_4^Q(\textbf{q})\right)^{\rm T} k_4^M(\textbf{p}),
\end{equation}
\begin{equation}\label{eq:soft}
 W_4(\textbf{q},\textbf{p}) = \text{SoftMax}_\textbf{p}(s_4(\textbf{q},\textbf{p})),
\end{equation}
where $\textbf{p}=\left( p_t,p_x,p_y \right) $ and $\textbf{q}= \left( q_x,q_y \right) $ denote the grid cell locations in the memory and query, respectively, and thus $\left(k_4^Q(\textbf{q})\right)^{\rm T} k_4^M(\textbf{p})$ performs the dot product between two $\frac{C_4}{8}$-dimensional vectors at the locations $\textbf{p}$ and $\textbf{q}$ in the memory and query, and ${\rm{T}}$ indicates the transpose operator. $s_4\in\mathbb{R}^{H_4W_4 \times T_4H_4 W_4 }$ thus contains similarity values in every space-time locations, and $\text{SoftMax}_\textbf{p}$ performs the SoftMax operation along the memory axis. $v_4^M$ is multiplied with $W_4$ and then concatenated with $v_4^Q$ as follows:
\begin{equation}\label{eq:y4}
  y_4 = \left[v^Q_4, v_4^M W_4^{\rm T} \right], 
\end{equation}
where [,] represents the concatenation along the feature dimension. $y_4\in\mathbb{R}^{{C_4} \times H_4 W_4}$ represents the output of the memory read operation at the fourth stage.

Since the computational complexity required for (Eq. \ref{eq:s4})-(Eq. \ref{eq:y4}) grows quadratically with respect to the size of the feature map, we apply an efficient read operation called top-$k$ read~\cite{seong2021hierarchical, cheng2021modular} at the earlier stages. Specifically, the affinity maps for the earlier stages $s_i$ $(i=1,2,3)$ are obtained using only the top-$k$ indices as follows:
\begin{equation}\label{eq:s321}
  s_i(\textbf{q}, :) = \left(k_i^Q(\textbf{q})\right)^{\rm T} k_i^M(\textbf{p}), {\textbf{p}} \in \Omega^i_\textbf{q} ,
\end{equation}
where $\Omega^i_\textbf{q}$ denotes the set of the top-$k$ indices for the query pixel $\textbf{q}$ found from $s_4$ that are mapped to the $i$-th stage~\cite{seong2021hierarchical}. $\Omega^3_\textbf{q}$, $\Omega^2_\textbf{q}$, and $\Omega^1_\textbf{q}$ contain 4$k$ positions in $k_3^M$, 16$k$ positions in $k_2^M$, and 64$k$ positions in $k_1^M$, respectively, such that more pixels can be matched at the higher scales. $s_i(\textbf{q}, :)$ thus collects the similarity values in these top $4^{4-i}k$ locations.
$W_i\in\mathbb{R}^{H_iW_i \times 4^{4-i}k}$ is obtained by applying the SoftMax operation to $s_i$. Finally, only a sparse matching to the selected locations from the memory is performed as follows:
\begin{equation}\label{eq:y321}
   y_i = \left[v^Q_i, \tilde{v}_i^M W_i^{\rm T}\right], i=\{1,2,3\},
\end{equation}
where $\tilde{v}_i^M\in\mathbb{R}^{\frac{C_i}{2} \times 4^{4-i}k }$ is constructed by sampling $4^{4-i}k$ samples for each query pixel from ${v}_i^M$. The output of the memory read $y_i\in\mathbb{R}^{{C_i} \times H_i W_i}$ is passed to the decoder to extract a mask prediction.


\subsubsection{Decoder}
We use the refinement module in \cite{8578868} as the building block of our decoder. The output of the last stage memory read, i.e., $y_4$, is gradually upsampled with convolutional layers. The refinement module at each stage also takes the output of the top-$k$ memory read at the corresponding scale through skip connections. The refinement module produces an object mask with the size $H_1 \times W_1$ $\left(= \frac{H}{4} \times \frac{W}{4} \right)$, which is bilinearly upsampled to the original resolution. The soft aggregation of the output masks~\cite{oh2019video} is applied when handling multiple objects.

\subsection{Architecture Variants}
We introduce four architecture variants of HST, i.e., HST-T, HST-S, HST-B, and HST-L, by using the following hyper-parameter settings.
\begin{itemize}

    \item HST-T: $C=96$, layer numbers = $\{2,2,6,2\}$, window size = 7
    \item HST-S: $C=96$, layer numbers = $\{2,2,18,2\}$, window size = 7
    \item HST-B: $C=128$, layer numbers = $\{2,2,18,2\}$, window size = 12
    \item HST-L: $C=192$, layer numbers = $\{2,2,18,2\}$, window size = 12

\end{itemize}
The image and video Transformers of the base model (HST-B) require 193.6 M parameters, and those for the rest three variants require approximately 0.25$\times$ (HST-T), 0.5$\times$ (HST-S), and 2$\times$ (HST-L) of the parameters, respectively. 


\section{Experiments}

\subsection{Implementation Details}
\noindent \textbf{Training.}
We followed the same training strategy as STM~\cite{oh2019video}, HMMN~\cite{seong2021hierarchical}, PCVOS~\cite{park2022per}. We initialized the image Transformer blocks with ImageNet pre-trained weights and randomly initialized the other layers. Because the Video Transformer blocks take additional masks as input, they cannot be simply pre-trained using video datasets. Therefore, we initialized the Video Transformer blocks by replicating the image Transformer block's ImageNet pre-trained weights along the temporal dimension. Then, we pre-trained HST on the image datasets, including MSRA10K, ECSSD, PASCAL-S, PASCAL VOC2012, and COCO datasets~\cite{cheng2014global,shi2015hierarchical,jiang2013salient,everingham2010pascal,lin2014microsoft}. For these image datasets, we synthesized three consecutive frames by augmenting each image via random affine transformations, including rotation, shearing, zooming, translation, and cropping. 

\begin{table}[t]
\begin{center}
\caption{Comparison on the DAVIS 2016 validation set. (+Y) indicates YouTube-VOS is additionally used for training, and OL denotes the use of online-learning strategies during test time. * denotes time measurements from the corresponding papers. $^{\dag}$ denotes the results obtained using the first and previous frames as input of the video Transformer. }
\label{table:DAVIS2016}

\setlength{\tabcolsep}{6pt}
    \begin{tabular}{lccccc}
    \toprule
    Method    & OL           & $\mathcal{J}\&\mathcal{F}$ & $\mathcal{J}$ & $\mathcal{F}$   & Time (s) \\
    \midrule
    OSVOS~\cite{caelles2017one}   &\checkmark    & 80.2 & 79.8 & 80.6  & 9*       \\
    MaskRNN~\cite{hu2017maskrnn}   &\checkmark    & 80.8 & 80.7 & 80.9  &  -       \\
    VideoMatch~\cite{hu2018videomatch} &             &   -  & 81.0 &   -   & 0.32*    \\
    FEELVOS~\cite{voigtlaender2019feelvos} (+Y)&            & 81.7 & 81.1 & 82.2  & 0.45*    \\
    PReMVOS~\cite{luiten2018premvos}   &\checkmark    & 86.8 & 84.9  & 88.6  & 30*     \\
    STM~\cite{oh2019video} (+Y)  &              & 89.3 & 88.7  & 89.9  & 0.10  \\
    CFBI~\cite{yang2020collaborative} (+Y) &              & 89.4 & 88.3  & 90.5  & 0.13  \\
    KMN~\cite{seong2020kernelized} (+Y)  &              & 90.5 & 89.5  & 91.5  & -  \\
    HMMN~\cite{seong2021hierarchical} (+Y) &              & 90.4 & 89.6  & 92.0  & 0.07  \\
    SITVOS~\cite{lan2022siamese} (+Y) &       & 90.5 & 89.5 & 91.4 & 0.09 \\
    AOT~\cite{yang2021associating} (+Y) & & 91.1 & 90.1 & 92.1& 0.06 \\
    MaskVOS~\cite{wang2022delving} (+Y) &  & 91.1 & 89.9 & 92.3 &0.11\\
    STCN~\cite{cheng2021rethinking} (+Y) & & 91.6 & 90.8 & 92.5& 0.05 \\
    AOCVOS~\cite{xu2022towards} (+Y) &  & 91.6 & 88.5 & 94.7 & 0.32\\
    PCVOS~\cite{park2022per} (+Y) & &  91.9 & 90.8 & 93.0 & 0.11 \\
    QDMN~\cite{liu2022learning} (+Y) & & 92.0 & 90.7 & 93.2 & 0.13 \\
    \midrule
    HST-T$^{\dag}$ (+Y)   &     & 92.1 & 91.0 & 93.1 & 0.11        \\
    HST-T (+Y) & & 92.9 & 92.6 & 93.2 & 0.21 \\

    HST-S$^{\dag}$ (+Y)   &     & 92.2 & 91.2 & 93.1 & 0.15         \\
    HST-S (+Y) & & 93.0 & 92.2 & 93.8 &0.28\\
    
    HST-B$^{\dag}$ (+Y)   &              & 93.1 & 91.9  & 94.3  & 0.24  \\
    HST-B (+Y)   &              & 94.0 & 93.2  & 94.8  & 0.36  \\

    HST-L$^{\dag}$ (+Y)   &   & 93.7 & 92.8 & 94.5 & 0.29             \\
    HST-L (+Y) & & \textbf{94.2} & \textbf{93.4} & \textbf{95.0} & 0.51 \\
    \bottomrule
    \end{tabular}
\end{center}
\end{table}

After the pre-training on the synthesized image dataset, the main training was conducted using either DAVIS 2017 or YouTube-VOS 2019 training set, depending on the target benchmark. During the main training, three frames were randomly sampled from a video with a gradually increasing maximum interval (from 0 to 25). During both the pre-training and main training, we minimized the pixel-wise cross-entropy loss with Adam optimizer~\cite{kingma2014adam}, and the learning rate was set to 1e-5. 
We used an input size of 384 $\times$ 384 and set $P_x$ = 4, $P_y$ = 4, and $P_t$ = 1. Following STM, we employed the soft aggregation when multiple target objects exist in a video~\cite{oh2019video}.
\begin{table}[t]
\setlength{\tabcolsep}{14pt}
\begin{center}
\caption{Comparison on the DAVIS 2017 validation and test-dev set. (+Y) indicates YouTube-VOS is additionally used for training.}
\label{table:DAVIS2017}
\begin{tabular}{l c c c}
        \toprule[1pt]
        Methods  & $\mathcal{J}\&\mathcal{F}$ & $\mathcal{J}$ & $\mathcal{F}$ \\
        \midrule[0.5pt]
        \multicolumn{4}{c}{\textit{Validation 2017 Split}} \\
        \midrule[0.5pt]
        STM~\cite{oh2019video}  &               71.6 & 69.2  & 74.0 \\
        STM~\cite{oh2019video} (+Y)  &               81.8 & 79.2  & 84.3 \\
        CFBI~\cite{yang2020collaborative} &               74.9 & 72.1  & 77.7 \\
        CFBI~\cite{yang2020collaborative} (+Y) &               81.9 & 79.1  & 84.6 \\
        SST~\cite{duke2021sstvos}  & 78.4 & 75.4 & 81.4\\
        SST~\cite{duke2021sstvos} (+Y) & 82.5 & 79.9 & 85.1\\
        KMN~\cite{seong2020kernelized}   &               76.0 & 74.2  & 77.8 \\
        KMN~\cite{seong2020kernelized} (+Y)  &               82.8 & 80.0  & 85.6 \\
        CFBI+~\cite{yang2021collaborative} (+Y) & 82.9 & 80.1 & 85.7 \\
        RMNet~\cite{xie2021efficient} (+Y) & 83.5 & 81.0 & 86.0 \\
        SITVOS~\cite{lan2022siamese} (+Y) & 83.5 & 80.4 & 86.5 \\
        AOCVOS~\cite{xu2022towards} (+Y) & 83.8 & 81.7 & 85.9 \\
        HMMN~\cite{seong2021hierarchical} (+Y) &               84.7 & 81.9  & 87.5 \\
        AOT~\cite{yang2021associating}  &  79.3 & 76.5 & 82.2 \\
        AOT~\cite{yang2021associating} (+Y) &  84.9 & 82.3 & 87.5 \\
        STCN\cite{cheng2021rethinking} (+Y) & 85.4 &82.6 & 88.6 \\
        MaskVOS~\cite{wang2022delving} (+Y) & 85.5 & 82.0 & 89.0 \\
        QDMN~\cite{liu2022learning} (+Y) & 85.6 & 82.5 & 88.6 \\
        PCVOS~\cite{park2022per} (+Y) & \textbf{86.1} & \textbf{83.0} & 89.2 \\
        \midrule[0.5pt]
        HST-T (+Y) &  83.6 & 80.9 & 86.2\\
        HST-S (+Y) &  84.0 & 80.7 & 87.3\\
        HST-B  &  79.9 & 76.9 & 82.9\\
        HST-B (+Y)   &               85.9 & 82.5  & \textbf{89.2} \\ 
        HST-L (+Y) &  85.6 & 82.2 & 89.0 \\
        \midrule[0.5pt] 
         \multicolumn{4}{c}{\textit{Testing 2017 Split}} \\
         \midrule[0.5pt]
             STM~\cite{oh2019video} (+Y)  &               72.2 & 69.3  & 75.2 \\
             CFBI~\cite{yang2020collaborative} (+Y) &               74.8 & 71.1  & 78.5 \\
             KMN~\cite{seong2020kernelized} (+Y)  &               77.2 & 74.1  & 80.3 \\
             CFBI+~\cite{yang2021collaborative}] (+Y) & 78.0 & 74.4 & 81.6 \\
             HMMN~\cite{seong2021hierarchical} (+Y) &               78.6 & 74.7  & 82.5 \\   
             STCN~\cite{cheng2021rethinking} (+Y) &  77.8 & 74.3 & 81.3 \\
             AOCVOS~\cite{xu2022towards} (+Y) & 79.3 & 74.7 & 83.9\\
             AOT~\cite{yang2021associating} (+Y) &  79.6 & 75.9 & 83.3\\
             \midrule[0.5pt] 
             HST-T (+Y) &  78.9 & 75.7 & 82.2 \\
             HST-S (+Y) &  79.2 & 75.8 & 82.6 \\
             HST-B (+Y)   &  79.9 & 76.5  & 83.4\\
             HST-L (+Y) & \textbf{80.2} & \textbf{76.8} & \textbf{83.6} \\
        \bottomrule[1pt]
        \end{tabular}
    \end{center}
\end{table}

\noindent \textbf{Inference.}
We used the first, previous, and intermediate frames sampled at every eight frames as input for the video Transformer. We used the same number of $k$ = 128 for top-$k$ guided memory matching during the training and inference. We measured the run-time of our and compared methods using two NVIDIA RTX 3090 GPUs.

\subsection{Comparisons}

\begin{table}[t]
\begin{center}
\caption{Quantitative evaluation on the YouTube-VOS validation set}
\label{table:YouTube}
\setlength{\tabcolsep}{7pt}
\begin{tabular}{lccccc}
        \toprule[1pt]
                    &   &  \multicolumn{2}{c}{Seen}  &    \multicolumn{2}{c}{Unseen}   \\
        \midrule[0.5pt]
         Methods & $\mathcal{J}\&\mathcal{F}$ & $\mathcal{J_S}$ & $\mathcal{F_S}$ & $\mathcal{J_U}$ & $\mathcal{F_U}$  \\
        \midrule[0.5pt]
        \multicolumn{6}{c}{\textit{Validation 2018 Split}} \\
        \midrule[0.5pt]
        STM~\cite{oh2019video}   &  79.4  &  79.7  &  84.2  &  72.8  &  80.9 \\
        SITVOS~\cite{lan2022siamese} & 81.3 & 79.9 & 84.3 & 76.4 & 84.4 \\
        KMN~\cite{seong2020kernelized}  &  81.4  &  81.4  &  85.6  &  75.3  &  83.3\\
        CFBI~\cite{yang2020collaborative} &  81.4  &  81.1  & 85.8  & 75.3  & 83.4 \\
        SST~\cite{duke2021sstvos} & 81.7  &  81.2  &  -  &  76.0  &  -  \\
        MaskVOS~\cite{wang2022delving} & 81.9 & 81.4 & 86.6 & 75.9 & 83.9 \\
        CFBI+~\cite{yang2021collaborative} &  82.8  &  81.8  & 86.6  & 77.1  & 85.6  \\
        
            HMMN~\cite{seong2021hierarchical} & 82.6 & 82.1 & 87.0 & 76.8 & 84.6 \\
            STCN~\cite{cheng2021rethinking} & 83.0 & 81.9 & 86.5 & 77.9 & 85.7\\
            QDMN~\cite{liu2022learning} & 83.8 & 82.7 & 87.5 & 78.4 & 86.4 \\
            AOCVOS~\cite{xu2022towards} & 84.0 & 83.2 & 87.8 & 79.3 & 87.3\\
            AOT~\cite{yang2021associating} & 84.1 & 83.7 & 88.5 & 78.1 & 86.1 \\
            PCVOS~\cite{park2022per} & 84.6 & 83.0 & 88.0 & 79.6 & 87.9 \\
            
            \midrule[0.5pt] 
            HST-T & 83.2 & 82.7 & 86.8 & 78.2 & 85.1\\
            HST-S  & 83.9 & 83.4 & 87.0 & 78.4 & 86.8\\
            HST-B & 85.0    & 84.3 & \textbf{89.2} & 79.0 & 87.6        \\ 
            HST-L & \textbf{85.1} & \textbf{84.4} & 89.1 & \textbf{79.2} & \textbf{87.8} \\
        \midrule[0.5pt]
        \multicolumn{6}{c}{\textit{Validation 2019 Split}} \\
        \midrule[0.5pt]
        CFBI~\cite{yang2020collaborative}  &  81.0  &  {80.6}  & {85.1}  & {75.2}  & {83.0}   \\
        SST~\cite{duke2021sstvos} & 81.8  &  80.9  &  -  &  76.6  &  -  \\
        CFBI+~\cite{yang2021collaborative} &  82.6  &  81.7  & 86.2  & 77.1  & 85.2 \\
        
            HMMN~\cite{seong2021hierarchical} & 82.6 & 82.1 & 87.0 & 77.3 & 85.0 \\
            STCN~\cite{cheng2021rethinking} & 82.7 & 81.1 & 85.4 & 78.2 & 85.9 \\
            AOT~\cite{yang2021associating} & 84.1 & 83.5 & 88.1 & 78.4 & 86.3 \\
            AOCVOS~\cite{xu2022towards} & 84.1 & 82.7 & 87.1 & 80.0 & 87.8 \\
            PCVOS~\cite{park2022per} & 84.6 & 82.6 & 87.3 & 80.0 & 88.3\\
        \midrule[0.5pt] 
            HST-T  & 83.5 & 82.9 & 87.4 & 78.2 & 85.5 \\
            HST-S  & 84.1 & 83.3 & 88.3 & 78.0 & 86.7 \\
            HST-B   & 84.9    & 83.6 & \textbf{88.5} & 79.5 & 88.1   \\
            HST-L  & \textbf{85.0} & \textbf{83.7} & 88.3 & \textbf{79.7} & \textbf{88.3} \\
        \bottomrule[1pt]
        \end{tabular}

    \end{center}
\end{table}

We compared our HST with state-of-the-art methods on the DAVIS~\cite{perazzi2016benchmark,pont20172017} and YouTube-VOS~\cite{xu2018youtube} benchmarks.
For the DAVIS benchmark, 60 videos from the DAVIS 2017 training set were used for the main training, following the standard protocol. In addition, we report our results on the DAVIS benchmark using additional training videos from Youtube-VOS for a fair comparison with several recent methods. For the Youtube-VOS benchmark, 3471 videos in the training set were used for training. 

\begin{table}[t]
\begin{center}
\caption{Ablation studies for HST-B on the DAVIS 2017 validation set. DM: Dense matching}
\setlength{\tabcolsep}{4pt}
\label{tab:ablation}
\begin{tabular}{llccc}
\toprule[1pt]
Ablation&Method&        $\mathcal{J}\&\mathcal{F}$ & $\mathcal{J}$ & $\mathcal{F}$  \\
\midrule[0.5pt]
\multicolumn{5}{c}{1.~Effect of pre-training} \\     
\midrule[0.5pt]
{\multirow{3}{*}{Training}} & \multicolumn{1}{|l}{Pre.}  &  74.2   & 70.4 &  76.3  \\
                          & \multicolumn{1}{|l}{Main} & 82.8 & 80.0 & 85.6 \\
                          & \multicolumn{1}{|l}{Full} & 85.9 & 82.5 &  89.2 \\
\midrule[0.5pt]   
\multicolumn{5}{c}{2.~Comparison of memory management strategies} \\     
\midrule[0.5pt]
{\multirow{2}{*}{Memory frames}} & \multicolumn{1}{|l}{First \& prev.}  &  84.9   & 81.6 & 88.2   \\
                          & \multicolumn{1}{|l}{+ Every 8 frames} & 85.9 & 82.5 &  89.2 \\
\midrule[0.5pt]   
\multicolumn{5}{c}{3.~Effect on hierarchical memory read} \\     
\midrule[0.5pt]
{\multirow{3}{*}{Memory read}} & \multicolumn{1}{|l}{Last stage only}  & 83.5    & 80.3 & 86.7   \\
                          & \multicolumn{1}{|l}{All stages w/ top-$k$} & 85.9 & 82.5 &  89.2\\
                          & \multicolumn{1}{|l}{All stages w/ DM} & 86.4 & 83.6  &  89.7  \\
\midrule[0.5pt]  
\multicolumn{5}{c}{4.~Effect on utilization of other object masks} \\     
\midrule[0.5pt]
{\multirow{2}{*}{Mask}} & \multicolumn{1}{|l}{w/o other object mask} & 84.1 & 81.2 & 86.9  \\
                          & \multicolumn{1}{|l}{w/ other object mask} & 85.9 & 82.5 &  89.2  \\
\midrule[0.5pt] 
\multicolumn{5}{c}{5.~Effect on spatiotemporal feature} \\     
\midrule[0.5pt]
{\multirow{2}{*}{Feature}} & \multicolumn{1}{|l}{Image feature only}& 83.0 & 79.9 & 86.1   \\
                          & \multicolumn{1}{|l}{Image and video features} & 85.9 & 82.5 &  89.2  \\
\bottomrule[1pt]
\end{tabular}
\end{center}
\end{table}

\begin{figure*}[ht]
  \centering
    \includegraphics[width=1.0\linewidth]{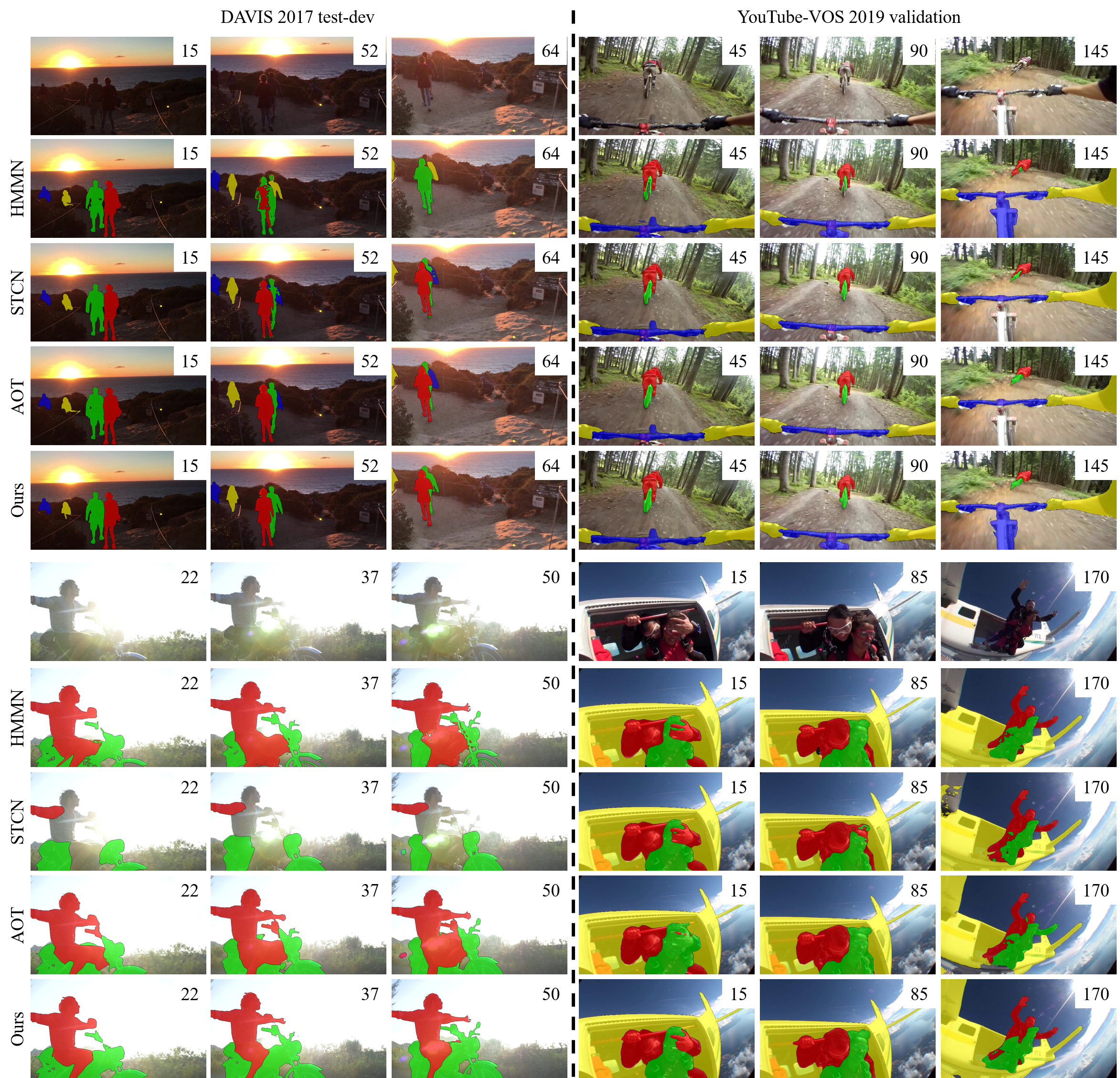}
  \caption{Qualitative performance comparison of HST with HMMN~\cite{seong2021hierarchical}, STCN~\cite{cheng2021rethinking}, and AOT~\cite{yang2021associating}.}
  \label{fig:result}
\end{figure*}
\begin{figure*}[htp]
  \centering
    \includegraphics[width=1\linewidth]{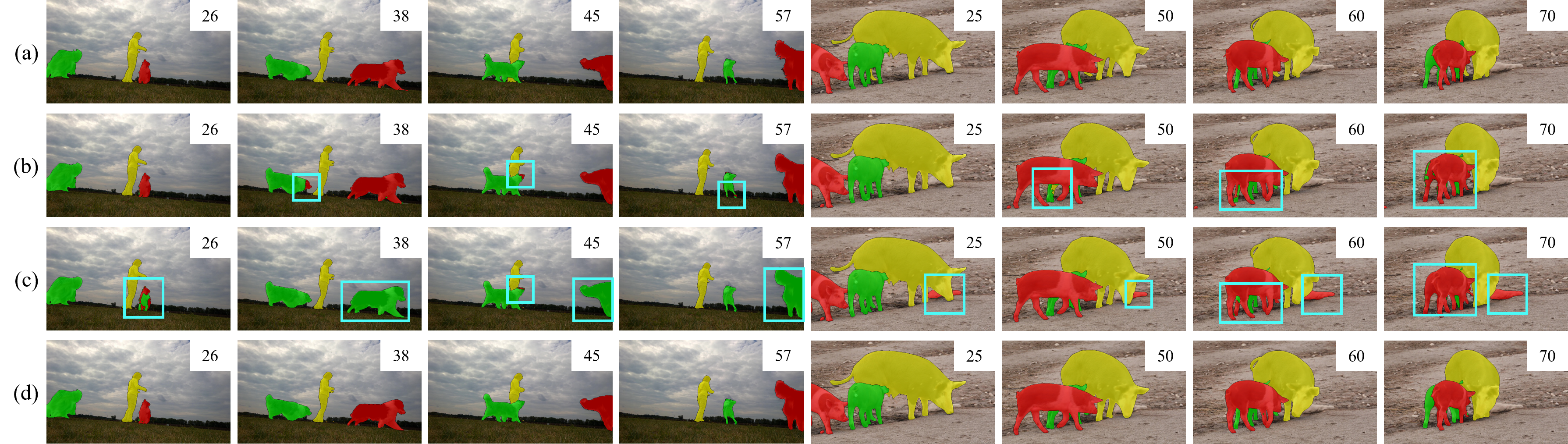}
  \caption{Qualitative comparison for the ablation studies: (a) Ground-truth, (b) result obtained w/o the other mask, (c) result obtained using image Transformer only, and (d) result of HST-B.}
  \label{fig:ablation_2}
\end{figure*}
DAVIS is a densely annotated VOS dataset and the most widely-used benchmark to evaluate VOS techniques. The DAVIS dataset consists of two sets: (1) DAVIS 2016, which is an object-level annotated dataset (single object); and (2) DAVIS 2017, which is an instance-level annotated dataset (multiple objects). The official metrics, i.e., region similarity $\mathcal{J}$ and contour accuracy $\mathcal{F}$, were measured for comparison. To evaluate HST, we used an input size of 480p resolution. As shown in Table~\ref{table:DAVIS2016}, HST-B outperforms the second-best method by $2.0\%$ $\mathcal{J}\&\mathcal{F}$ on the DAVIS 2016 validation set.  
Furthermore, additional experiments were conducted using the first and previous frames as input of the video Transformer to test the trade-off between the processing time and segmentation accuracy. We also conducted comparisons on the DAVIS 2017 validation and test-dev sets, and the results are given in Table~\ref{table:DAVIS2017}.
Our HST-B showed the competitive performance to PCVOS~\cite{park2022per} on the DAVIS 2017 validation set and achieved state-of-the-art performance on the DAVIS 2017 testing set.
Our HST-B trained without using the YouTube-VOS training dataset still showed improved performance over the other models trained without using the YouTube-VOS training dataset.

YouTube-VOS is a large-scale benchmark for VOS.
To evaluate our HST on the YouTube-VOS benchmark, we used an input size of 480p resolution. 
We measured the region similarity ($\mathcal{J_S}$ , $\mathcal{J_U}$ ) and contour accuracy ($\mathcal{F_U}$ , $\mathcal{F_U}$ ) for 65 seen and 26 unseen object categories separately.
Table~\ref{table:YouTube} shows the performance comparison of HST with state-of-the-art methods on the YouTube-VOS 2018 and 2019 validation sets, demonstrating that HST-B surpasses the state-of-the-art methods in both seen and unseen object categories.


Figure~\ref{fig:result} shows qualitative performance comparison with HMMN~\cite{seong2021hierarchical}, STCN~\cite{cheng2021rethinking}, and AOT~\cite{yang2021associating}. HMMN~\cite{seong2021hierarchical} failed in separating multiple occluded objects. STCN~\cite{cheng2021rethinking} and AOT~\cite{yang2021associating} produced incorrect results for incoming or outgoing objects in the scene.
On the other hand, HST predicted target objects accurately in these challenging scenarios. More results are provided in the supplementary material.

\subsection{Ablation Experiments}
We conducted ablation studies using HST-B on the DAVIS 2017 dataset. More details about the models used for the ablation studies are provided in the supplementary material.

\noindent \textbf{Pre-training.}
As shown in Table~$4.1$, the model pre-trained on the image datasets performed favorably with 74.2 $\%$ $\mathcal{J}\&\mathcal{F}$. Due to the effectiveness of the pre-training, the fully trained model exhibited 3.1 $\%$ higher $\mathcal{J}\&\mathcal{F}$ than the model trained on the DAVIS 2017 training dataset only.
Furthermore, it shows competitive performance without using synthesized static datasets.

\noindent \textbf{Memory management.}
As a default setting, HST uses the first, previous, and intermediate frames sampled at every eight frames as input for the video Transformer. Table~$4.2$ shows that HST performed reasonably well with 84.9 $\%$ $\mathcal{J}\&\mathcal{F}$ when only the first and previous frames were used as input. In environments where memory is scarce, it is advised to use only these two frames as input.

\noindent \textbf{Hierarchical memory read.} 
To show the effectiveness of using multi-scale features for the memory read, we obtained the result using the output of the memory read at the last stage only, i.e., $y_4$, as input for the decoder. As shown in Table~$4.3$, the performance decreased significantly, demonstrating the necessity of multi-scale features for precise mask decoding. In addition, when our hierarchical top-$k$ read was replaced by naive dense matching, we obtained a slightly better performance of 86.4 $\%$ $\mathcal{J}\&\mathcal{F}$. However, the dense matching at all stages required an average processing time of 2.78 s per frame, where the top-$k$ matching consumed 0.42 s per frame.


\noindent \textbf{Mask utilization.}
Our video Transformer takes given or predicted masks as input in addition to video frames. To better handle multiple object segmentation, we used the common strategy~\cite{oh2019video,seong2021hierarchical,seong2020kernelized,cheng2021rethinking} of including a binary object mask of other objects as additional input. Table~$4.4$ shows that the information on the other objects contributed to 1.8 $\%$ $\mathcal{J}\&\mathcal{F}$ improvement.

\noindent \textbf{Video Transformer.}
To demonstrate the effectiveness of using both image and video Transformers for spatiotemporal feature extraction, we built a model by applying only image Transformer to the current and past frames for feature extraction. Table $4.5$ shows that both image and video Transformers are essential for extracting spatiotemporal features, leading to 2.9 $\%$ $\mathcal{J}\&\mathcal{F}$ improvement.

Fig.~\ref{fig:ablation_2} shows some results for the ablation studies on the effect of other object masks and spatiotemporal features. As shown Fig.~\ref{fig:ablation_2}(b), the results obtained without other masks suffer from false matching due to similar appearances of the objects. Furthermore, as shown in Fig.~\ref{fig:ablation_2}(c), the results obtained using the image Transformer only show drifts over frames.


\section{Conclusions}
In this paper, we presented a novel VOS framework called HST that exploits image and video Transformers as a means of spatiotemporal feature extraction from a video. To take full advantage of image and video Transformers, we used image and video features as a query and memory, respectively, and matched them at multiple scales with efficient hierarchical memory read operations. HST showed state-of-the-art performance in several benchmarks, including the DAVIS 2016 and 2017 validation sets and YouTube-VOS 2018 and 2019 validation datasets. Considering the conciseness and technical advantages of HST, we hope our work can motivate future VOS studies.

\ifCLASSOPTIONcaptionsoff
  \newpage
\fi

\clearpage
\bibliographystyle{IEEEtran}
\bibliography{egbib}

\end{document}


\title{Hierarchical Spatiotemporal Transformers for Video Object Segmentation \\
\textit{-Supplementary Material-}}
\maketitle

In this \textit{supplementary material}, we provide detailed explanations for the models used in the ablation studies and additional results for visual comparison. 


\section{Details on Ablation Studies}

\paragraph{Mask utilization.}
\begin{figure}[h]
  \centering
    \includegraphics[width=\linewidth]{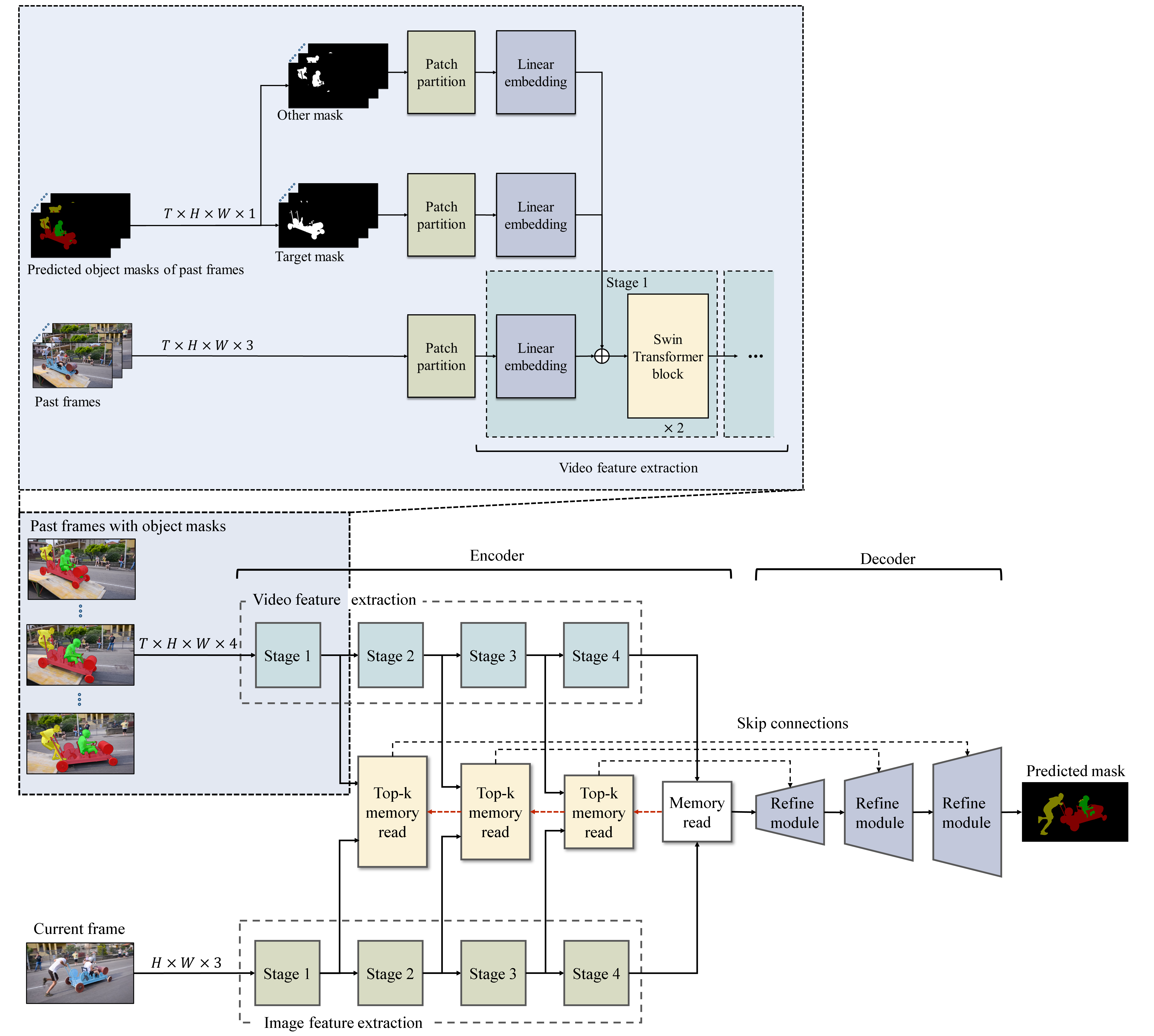}
  \caption{Detailed illustration of HST. Note that our video Transformer takes past frames and their predicted masks as input, where the predicted masks are further divided into the target object masks and other masks.}
  \label{fig:sup_mask util}
\end{figure}

\begin{figure*}[htp]
  \centering
    \includegraphics[width=0.9\linewidth]{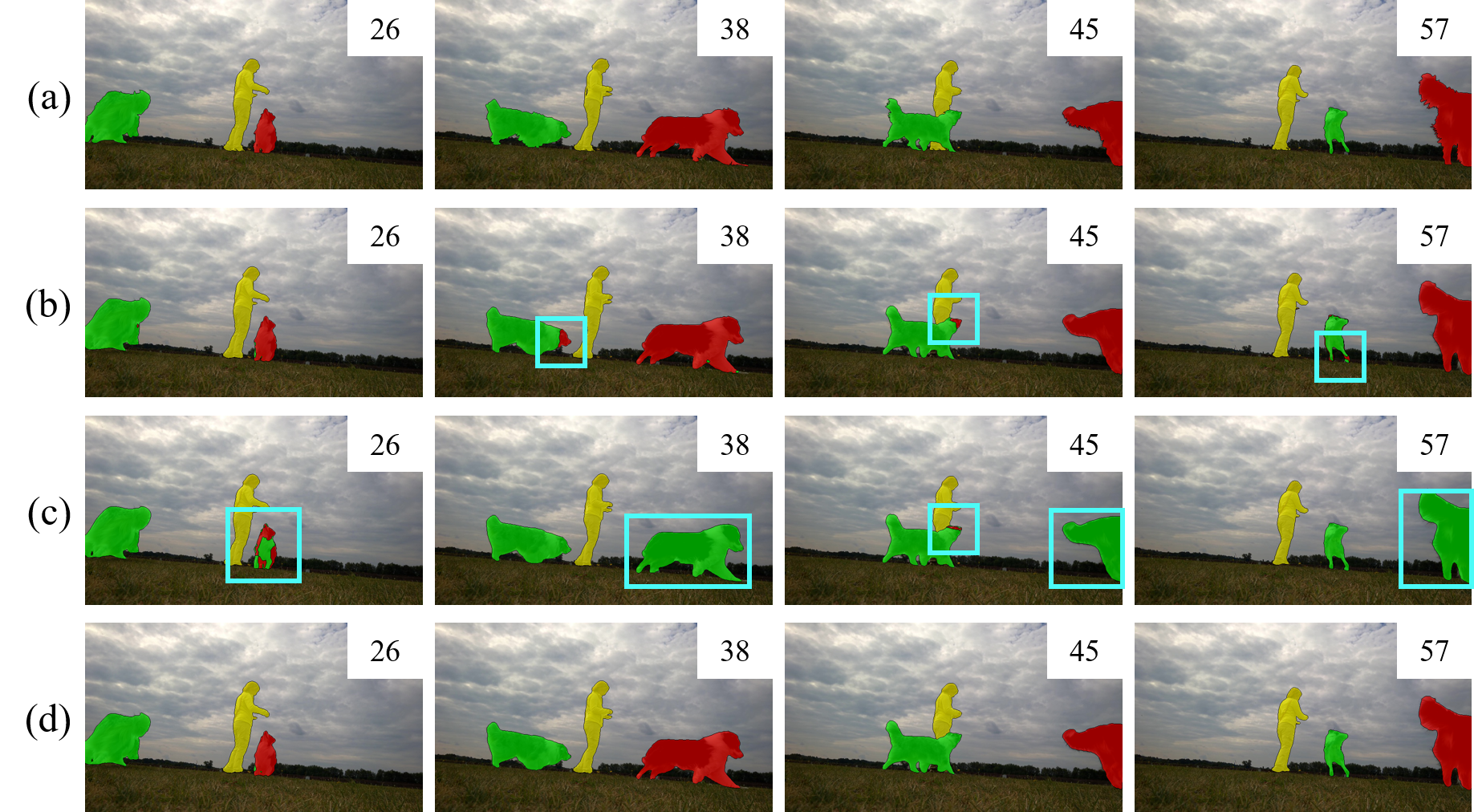}
  \caption{Qualitative comparison of ablation studies: (a) Ground-truth, (b) result obtained w/o the other mask, (c) result obtained using image Transformer only, and (d) result of HST-B.}
  \label{fig:sup_ablation_1}
\end{figure*}

\begin{figure*}[htp]
  \centering
    \includegraphics[width=0.9\linewidth]{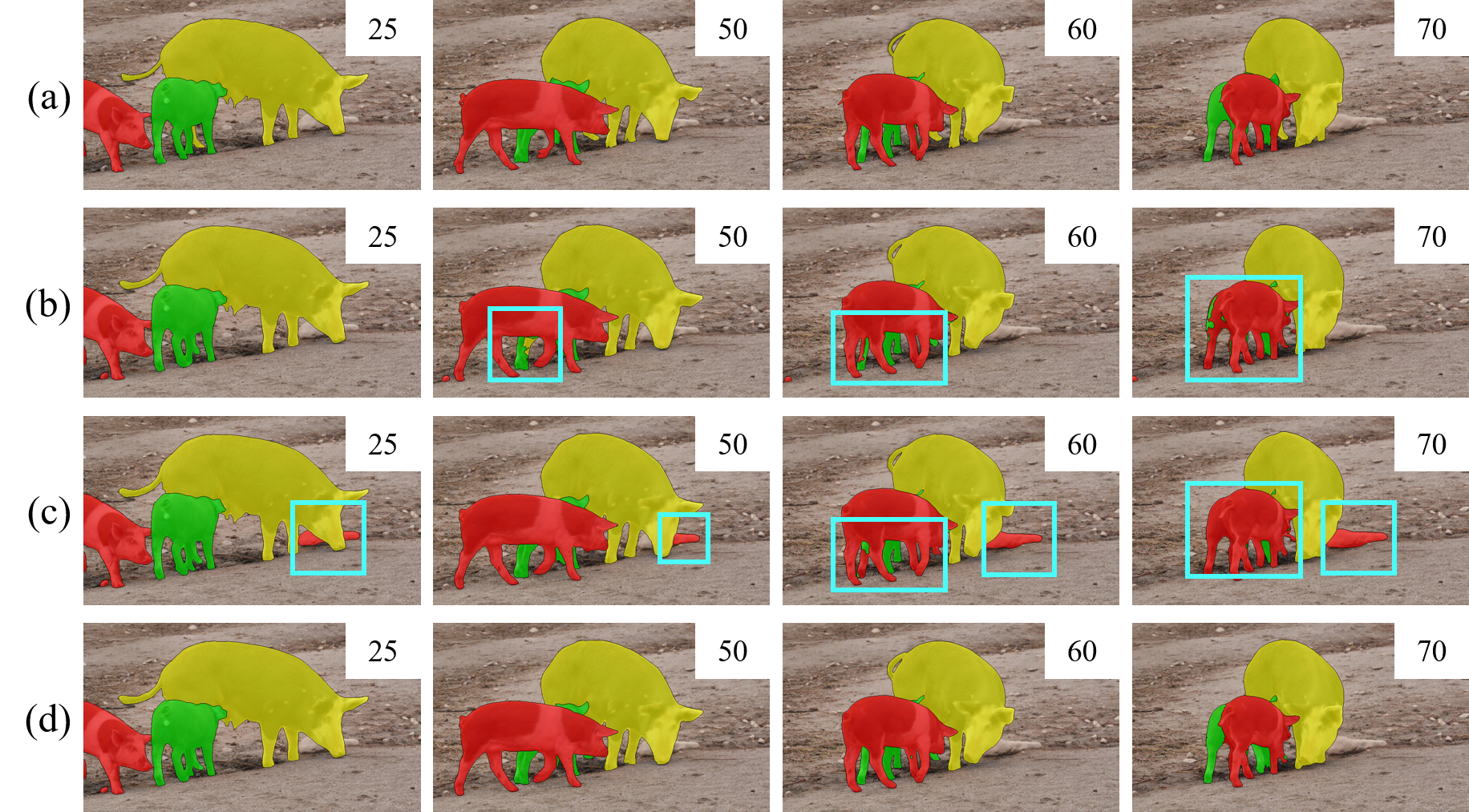}
  \caption{Qualitative comparison of ablation studies: (a) Ground-truth, (b) result obtained w/o the other mask, (c) result obtained using image Transformer only, and (d) result of HST-B.}
  \label{fig:sup_ablation_2}
\end{figure*}

The video Transformer of HST takes the predicted mask of the other objects, called `other mask,' as an additional input as shown in Figure~\ref{fig:sup_mask util}, following the common strategy~\cite{oh2019video,seong2021hierarchical,seong2020kernelized,cheng2021rethinking}. Note that the other mask is a single-channel binary map per frame (1: belongs to any other objects, 0: otherwise). Both the predicted mask of the target object and the other mask are fused into the image frame in the embedding space by passing through their respective linear embedding layers. This strategy can help HST find the target object outside the other objects as shown in Figures~ \ref{fig:sup_ablation_1} and~\ref{fig:sup_ablation_2}, leading to 1.8 $\%$ $\mathcal{J}\&\mathcal{F}$ improvement (See Table 4.4).

\paragraph{Video Transformer.}
\begin{figure}[h]
  \centering
    \includegraphics[width=0.98\linewidth]{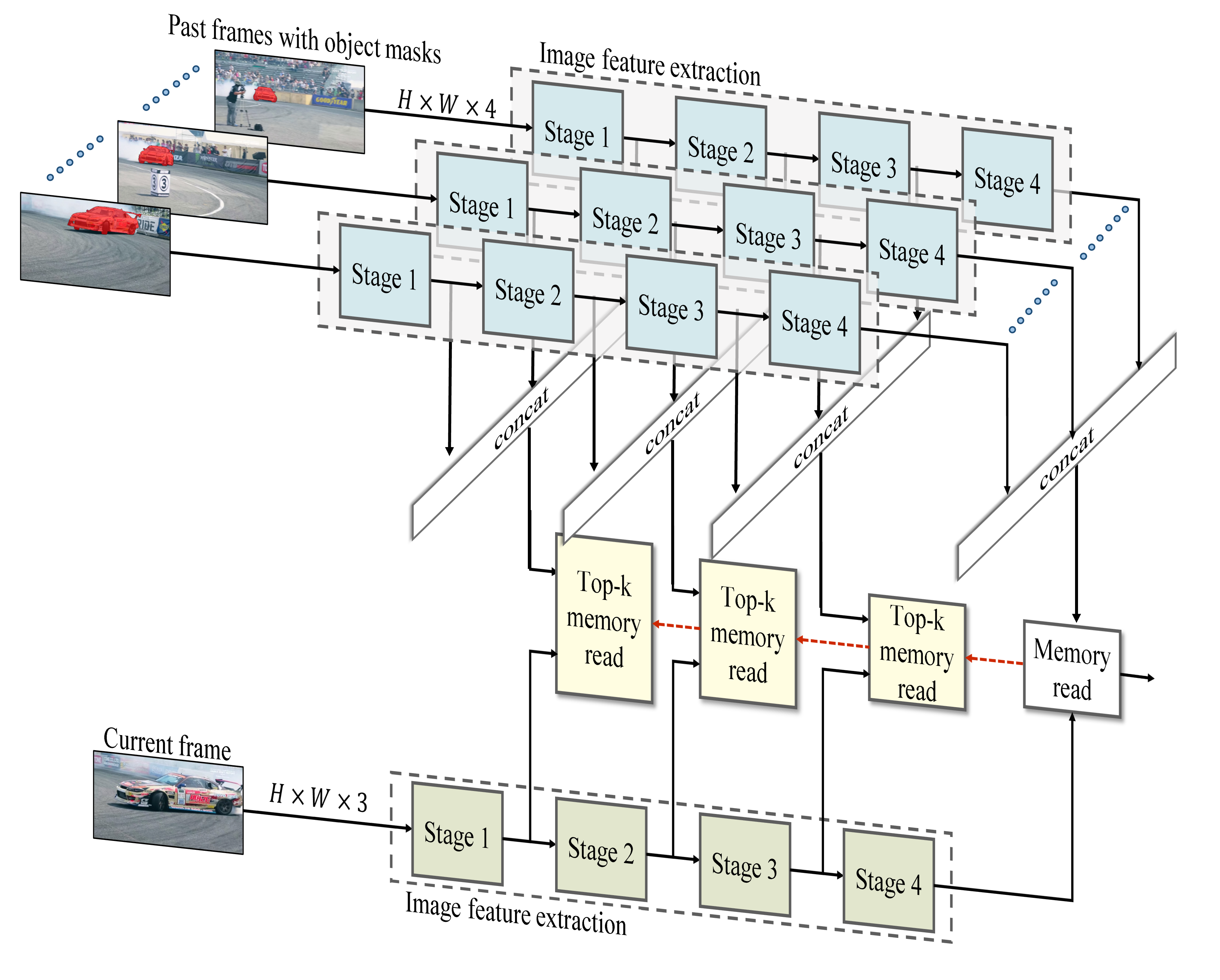}
  \caption{An encoder architecture embodied with image Transformer only. Note that the encoder performs the memory read operations using only image features extracted from current and past frames.}
  \label{fig:sup_image feature only}
\end{figure}
To demonstrate the effectiveness of using both image and video Transformers for spatiotemporal feature extraction, we built a model consisting of image Transformers only, as shown in Figure~\ref{fig:sup_image feature only}. Note that the video feature $F_{video}^{i}\in\mathbb{R}^{T_i\times H_i \times W_i \times C_i}$ at the $i$-th stage is constructed by concatenating the spatial features $F_{image}^{i} \in\mathbb{R}^{H_i \times W_i \times C_i}$ obtained from the past $T_i$ frames and their predicted object masks. In addition to the quantitative performance improvement that we demonstrated (See Table 4.5), we provide some results for visual comparison. As can be seen in Figures~\ref{fig:sup_ablation_1} and~\ref{fig:sup_ablation_2}, both image and video Transformers play an essential role in VOS, especially for these difficult scenes containing occluded and moving objects.


\paragraph{Hierarchical memory read.} 
In our implementation, HST processes multiple object segmentation in parallel, which is memory demanding but effective in reducing the processing time. However, in order to experiment with the model performing naive dense matching at all scales, we should process object segmentation sequentially to avoid memory overflow, leading to the average 2.78 s processing time per frame (See Table 4.3). 


\section{More Qualitative Results}
\begin{figure*}[htp]
  \centering
    \includegraphics[width=0.9\linewidth]{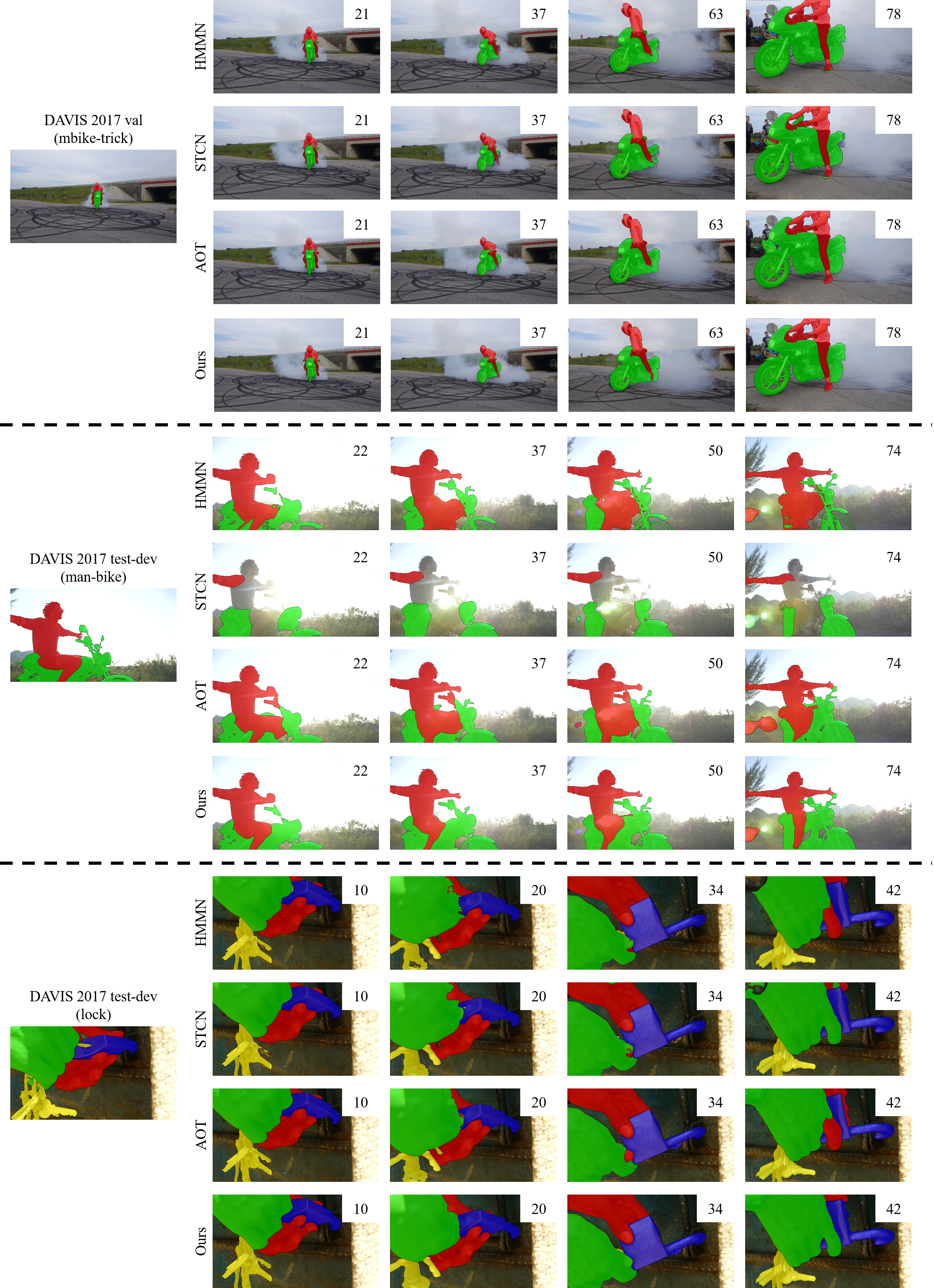}
  \caption{Qualitative performance comparison of HST with HMMN~\cite{seong2021hierarchical}, STCN~\cite{cheng2021rethinking}, and AOT~\cite{yang2021associating}.}
  \label{fig:sup_result}
\end{figure*}

In Figures~\ref{fig:sup_result}, we provide more results for visual comparison with the state-of-the-art methods: HMMN~\cite{seong2021hierarchical}, STCN~\cite{cheng2021rethinking}, and AOT~\cite{yang2021associating}. 



\clearpage
\bibliographystyle{IEEEtran}
\bibliography{egbib}